\def\year{2019}\relax
\documentclass[letterpaper]{article} %DO NOT CHANGE THIS
\usepackage{aaai19}  %Required
\usepackage{times}  %Required
\usepackage{helvet}  %Required
\usepackage{courier}  %Required
\usepackage{url}  %Required
\usepackage{graphicx}  %Required

\usepackage{amssymb}
\usepackage{amsmath,bm}
\usepackage{mathrsfs}
\usepackage{amsfonts}
\usepackage{multirow}
\usepackage{algorithm}
\usepackage{algorithmic}

\usepackage[shortlabels]{enumitem}

 \title{Learning Competitive and Discriminative Reconstructions for Anomaly Detection}
 \author{Kai Tian,\textsuperscript{\rm 1}
 	Shuigeng Zhou,\textsuperscript{\rm 1}\thanks{Correspondence author.}
 	Jianping Fan,\textsuperscript{\rm 2}
 	Jihong Guan\textsuperscript{\rm 3}\\
 	\textsuperscript{\rm 1}Shanghai Key Lab of Intelligent Information Processing, and School of Computer Science, Fudan University, China\\
 	\textsuperscript{\rm 2}Department of Computer Science, University of North Carolina at Charlotte, Charlotte, NC 28223 USA\\
 	\textsuperscript{\rm 3}Department of Computer Science \& Technology, Tongji University, China\\
 	\textsuperscript{\rm 1}\{ktian14, sgzhou\}@fudan.edu.cn; \textsuperscript{\rm 2}jfan@uncc.edu; \textsuperscript{\rm 3}jhguan@tongji.edu.cn\\
 }
 
\begin{document}

\maketitle
\begin{abstract}
Most of the existing methods for anomaly detection use only positive data to learn the data distribution, thus they usually need a pre-defined threshold at the detection stage to determine whether a test instance is an outlier. Unfortunately, a good threshold is vital for the performance and it is really hard to find an optimal one. In this paper, we take the discriminative information implied in unlabeled data into consideration and propose a new method for anomaly detection that can learn the labels of unlabelled data directly. Our proposed method has an end-to-end architecture with one encoder and two decoders that are trained to model inliers and outliers' data distributions in a competitive way. This architecture works in a discriminative manner without suffering from overfitting, and the training algorithm of our model is adopted from SGD, thus it  is efficient and scalable even for large-scale datasets. Empirical studies on 7 datasets including KDD99, MNIST, Caltech-256, and ImageNet etc. show that our model outperforms the state-of-the-art methods.
 \end{abstract}

\section{Introduction}
\label{Introduction}
Anomaly detection is to identify the data that do not conform to the expected normal patterns. These data may come from a new class or some noisy data that has no meaning. Usually, we call these abnormal data \emph{outliers}, and \emph{inliers} for the normal data. Anomaly detection is closely related with many real-world applications such as outlier detection, novelty detection in computer vision area ~\cite{khan2009survey,chandola2009anomaly,khan2014one} and medical diagnoses, drug discovery in bioinformatics~\cite{wei2018anomaly}. It can be categorized into one-class learning, where the profile of negative class is not well defined. According to the real application contexts, the negative data could be hard to collect or verify. Besides, there could be any kind of abnormal data that are unpredictable. Thus, those data are considered as novelties (or outliers), while the positive data (or inliers) are well characterized by the training data.
It is hard to use traditional multiple class classification methods to learn from only positive labeled data due to the inertness of classifiers.

Over the past decades, researchers have proposed lots of methods to deal with anomaly detection problems~\cite{eskin2000anomaly,chandola2009anomaly,liu2014unsupervised,malhotra2016lstm}. Generally, these methods either build a model configuration for normal data examples and identify the examples that disobey the normal profiles as outliers, or explicitly isolate outliers based on statistical or geometric measures of abnormality. Usually, different models have different capacities to characterize the data distribution. Most of the traditional methods are linear models which have limited model capacity. Although kernel function can be used to improve their capacities, it is not suitable for the context of high-dimensional and large-scale data.

Recently, deep learning methods have shown their powerful representation ability and gained immense success in many applications~\cite{vincent2010stacked,krizhevsky2012imagenet,bengio2013representation}. However, due to the unavailability of negative data, it is hard to  train a supervised deep neural network  for one-class classification directly. Although some efforts have been made to learn one-class classifier, most of them could not establish a discriminative model for anomaly detection. The detection is done by choosing a pre-defined threshold. From the probability perspective, it can be explained as that outliers should lie on the low-density regions of the model's distribution. However, since the outliers are not predictable, it is hard to determine a threshold that works for all cases. Meanwhile, as the model is only trained on the positive data, overfitting is another key factor that may destroy the model's generalization performance. That is the reason why we cannot simply train a DNN classifier based on positive data. Although one can use some strategies such as early stopping to avoid overfitting, it is very tricky and one can not decide when to stop is  best for the test data.

\begin{figure}[!htpb]
		\centering
		\includegraphics[width=0.5\textwidth]{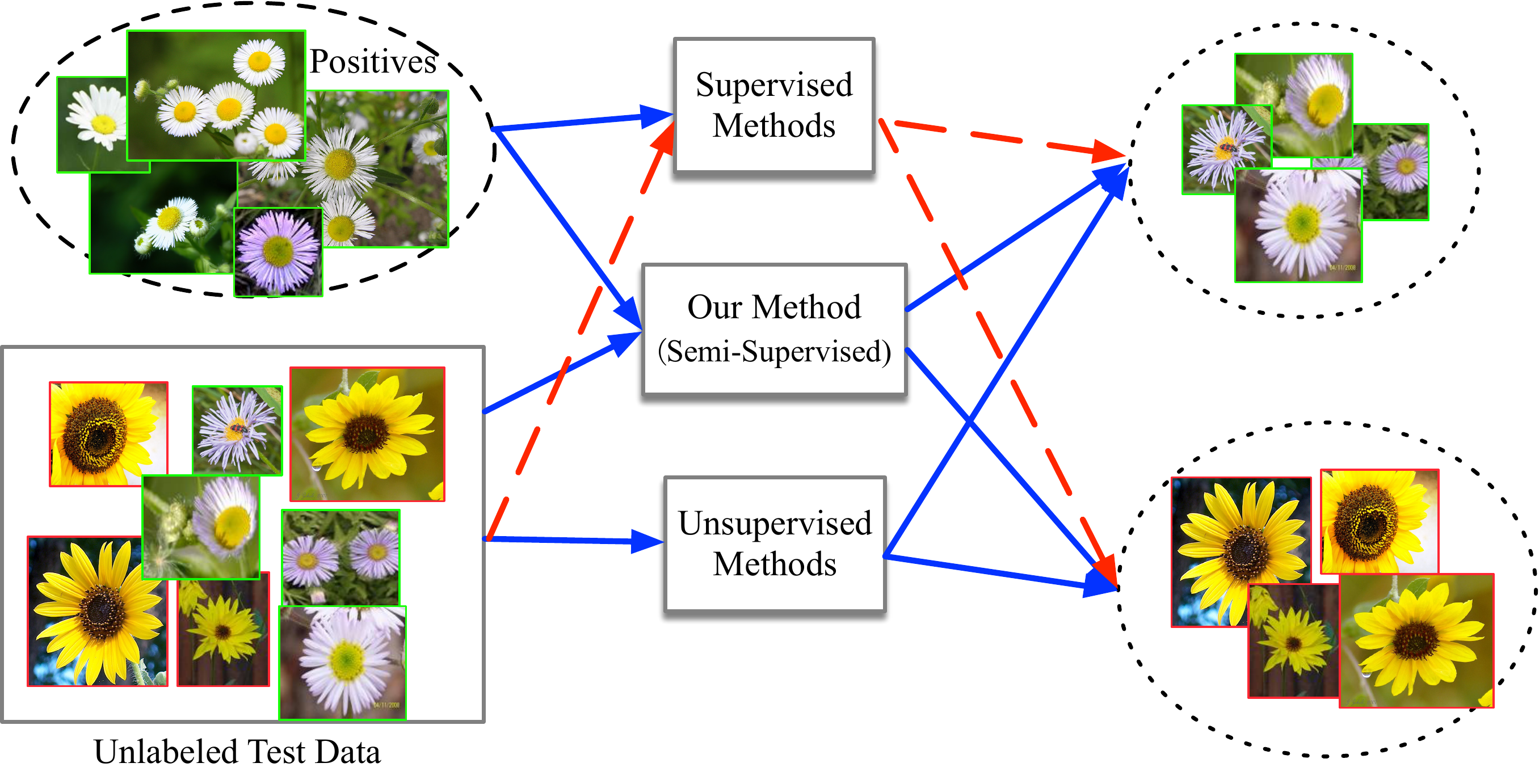}
	\caption{Comparison of the training/testing process among supervised methods, unsupervised methods and our method. The blue solid arrows indicate the training data flow. The red dashed arrows indicate the data flow of testing stage. Unsupervised methods and our method do not have a separate testing stage, they discriminate the outliers during the training stage.}
	\label{fig:demo}
\end{figure}

To address those issues, we propose a new model named \textit{Co}mpetitive \textit{R}econstruction \textit{A}utoencoder (\emph{CoRA}). 
Our model has the benefits of both supervised and unsupervised approaches. We formulate a transductive semi-supervised method for anomaly detection, which uses positive training data and unlabeled test data for learning.

Fig.~\ref{fig:demo} demonstrates the differences of the learning procedure between our method and most existing approaches. The proposed architecture, different from classical auto-encoder, comprises one encoder and two decoders.
These two decoders are designed to compete during the reconstruction procedure, one of them is learned to reconstruct positive data which is referred to as \emph{inlier decoder}, while the other is learned to reconstruct outliers which is referred to as \emph{outlier decoder}. With the guidance of  the positive training data, \emph{inlier decoder} can build a proper profile for positive class, while most of the anomaly data will be assigned to the \emph{outlier decoder}. Discriminative labeling is done by comparing the reconstruction errors of these two decoders. After training, the final assignments constitute the predictions for the unlabelled (or test) data.

It is known that auto-encoder will learn different feature subspaces for different data spaces. As our two competitive decoders share one encoder, it is reasonable to add a regularization term on the subspace to maximize the separability of  the  positive data and outliers, which will further improve \emph{CoRA}. Thus, in this paper we propose a manifold regularizer to preserve the data structures of positive data and  outliers in subspace.

In this paper, we propose a transductive semi-supervised anomaly detection approach that can be trained in an end-to-end manner. 
Our work is featured by the following merits:
\begin{itemize}[nosep]
	\item \textbf{Novel architecture.} A new transductive semi-supervised deep neural network framework implemented by an auto-encoder with one encoder and two competitive decoders is proposed for anomaly detection problems.
	\item \textbf{New criterion.} While the previous reconstruction-based algorithms use thresholds as the classification criterion, we propose a new scheme to make the decision, which can help us get rid of this hyper-parameter selection.
	\item \textbf{Robustness.} Extensive experiments show that the proposed model is more robust to the outlier ratio than many state-of-the-art methods.
	\item \textbf{Efficient and scalable optimization procedure.} We adopt the stochastic gradient descent (SGD) to our model, making it very efficient to train and can be used for large-scale data.
	\item \textbf{Comprehensive experiments.} We comprehensively evaluate the proposed model and compare it with a number of state-of-the-art methods on seven datasets.
\end{itemize}

\section{Related Work}
Anomaly detection belongs to one-class classification which is also closely related to outlier detection or novelty detection. The common objective of these applications is to discover novel concept, which has rarely or never shown up and is substantially different from the known concept.

\begin{table*}
	\caption{Comparison of state-of-the-art DNN-based models with our method from three perspectives.}
	\label{tb:diff}
	\begin{center}
		\begin{tabular}{cccc}
			
			\hline
			Method & Architecture & Criterion & Learning paradigm \\
			\hline
			DRAE~\cite{xia2015learning} & auto-encoder & Discriminative & Unsupervised\\
			DSEBM~\cite{zhai2016deep} & DNN & Threshold (energy) & Supervised \\
			DAOC~\cite{sabokrou2018adversarially} & GAN & Threshold (score) & Supervised\\
			SSGAN~\cite{kimura2018semi} & GAN & Threshold (score) & Inductive semi-supervised\\
			\emph{CoRA}&auto-encoder & Discriminative & Transductive semi-supervised \\
			\hline
		\end{tabular}
	\end{center}
\end{table*}
Conventional researches often model the positive/target class and reject the samples that do not following them. Methods in this category usually estimate probability density function from positive data. By building up a parametric or non-parametric probability estimator from the positive samples, including kernel density estimator (KDE)~\cite{parzen1962estimation} and more recent robust kernel density estimator (RKDE)~\cite{kim2012robust}, one can identify an outlier sample when it has low probability density. These methods take the assumption that positive data are more densely distributed. Other statistical models such as PCA~\cite{de2003framework,candes2011robust,xu2013outlier}
assume that positive data are more correlated with each other than outliers and they get better reconstruction from the low-dimensional subspace.

Most deep learning approaches for anomaly detection are built upon auto-encoders, where reconstruction error is used to discriminate outliers from inliers. Those methods learn from only positive data which induces some problems. On the one hand, these methods need to pre-define a threshold to determine the unlabelled samples, while it is hard to find an optimal threshold. On the other hand, training auto-encoders on positive data will suffer from the overfitting problem, especially when the number positive samples is small.

 Recently, ~\cite{zhai2016deep} proposed a new energy-based deep neural network to detect outliers. Instead of directly using the reconstruction error as decision criterion, they showed that energy could be another criterion for identifying outliers. However, this method still need a pre-defined threshold to discriminate outliers. Instead of autoenocders, some researchers tried to solve one-class classification problem with generative adversarial networks (GANs). One of them is to combine a denoising auto-encoder with a discriminator and train them in an adversarial way \cite{sabokrou2018adversarially}. The key idea is to enhance the decoder to reconstruct inliers perfectly while distort the outliers. Although they used discriminator of GAN to classify the samples, classification is actually based on a  task-dependent threshold, i.e. $x$ is a outlier if $f(x)<\sigma$.

Our method is closely related to some unsupervised approaches for anomaly detection or outlier removal. Different from supervised methods that just learn from positive data, unsupervised methods take the discriminative information from the unlabeled data by introducing  auxiliary labels, and maximize the discrimination between inliers and outliers, they iteratively refine the labels and finally output the predictions. An unsupervised
one-class learning (UOCL) method was proposed in ~\cite{liu2014unsupervised} by utilizing an auxiliary label variable, and jointly optimizing the kernel-based max-margin classifier and the soft-label assignment. Another deep learning based unsupervised approach is discriminative reconstruction auto-encoder (DRAE)~\cite{xia2015learning}, they also introduced discriminative labels for the unlabeled data and optimized the within-class variance of reconstruction error for both inliers and outliers in the learning process of auto-encoder.

Note that our formulation for anomaly detection is different from positive and unlabeled (PU) learning~\cite{elkan2008learning} or \emph{inductive} semi-supervised anomaly detection~\cite{kimura2018semi}. In both cases, it is assumed that the unlabeled data that used for training share the same distribution with the test data.
 In our case, similar to unsupervised approaches, only positive data and unlabelled (or test) data are used for training. Our method belongs to \emph{transductive} semi-supervised learning~\cite{chapelle2009semi} where learning is to infer the correct labels for the given unlabeled data only.

For better understanding the differences between most existing deep learning models and our method, we present a comparison of them from three perspectives in Table~\ref{tb:diff}.

\begin{figure}
	\centering
	\includegraphics[width=0.45\textwidth]{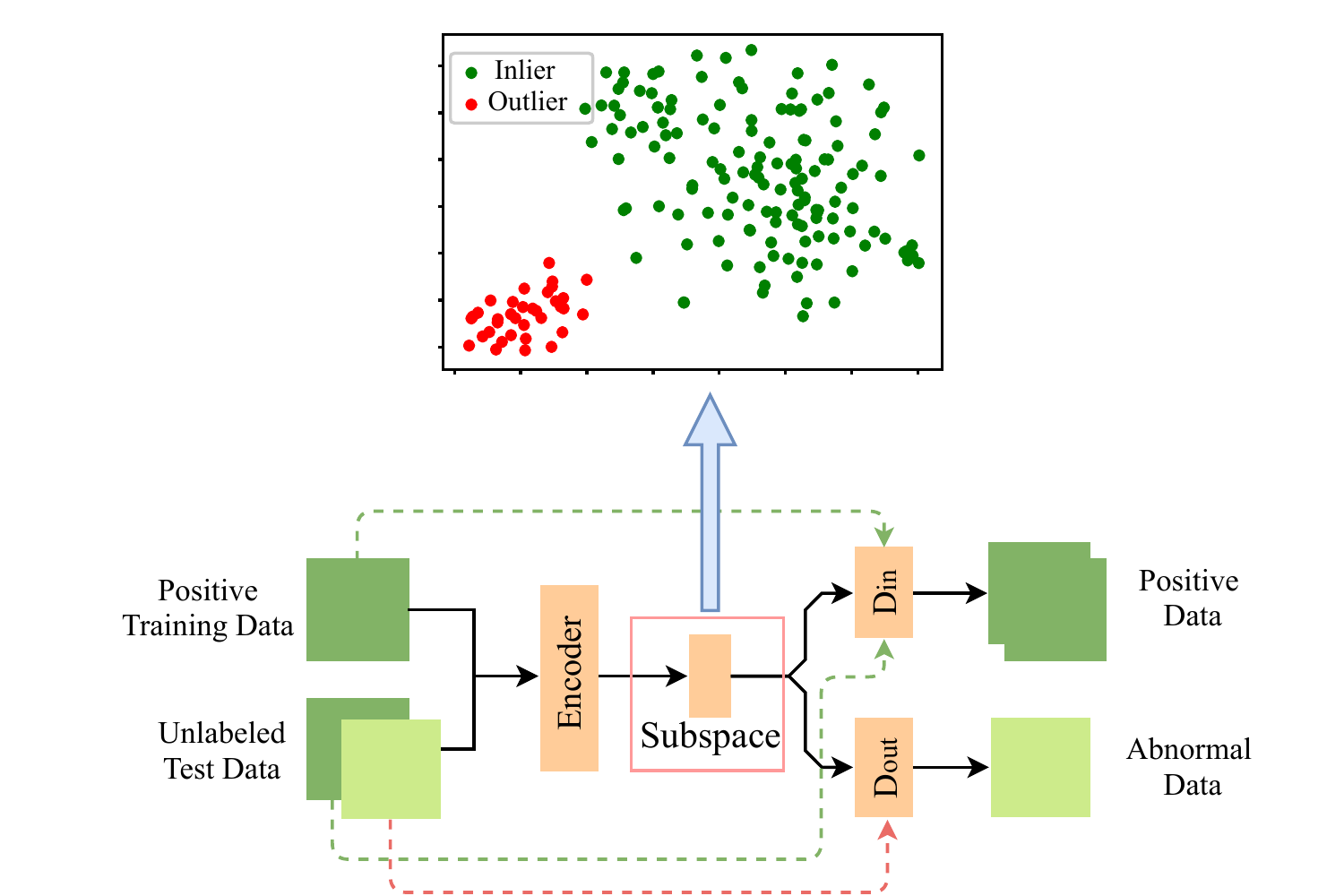}
	\caption{The architecture of the \emph{CoRA} model. Both training data and test data are used for learning. All data share the same encoder but have to choose one of the two different decoders. The encoder aims to map positive data and unlabelled data in the subspace separately. Green dotted lines indicates that positive data in both training and test datasets would be assigned to the \emph{inlier decoder}, while the red dotted line represents the assignment of abnormal data (or outliers).}
	\label{fig:arch}
\end{figure}

\section{Competitive Reconstruction Auto-encoder}
In this section, we present detailed description about our competitive reconstruction auto-encoder (\emph{CoRA}). We introduce the architecture and the objective function of \emph{CoRA} first and then we show how to optimize this model.

\subsection{Model Architecture}
The proposed competitive reconstruction auto-encoder is composed of three main modules: (1) encoder network $E$, (2) inlier decoder network $\mathcal{D}_{in}$, and (3) outlier decoder network $D_{out}$. The encoder maps the data into a common subspace shared by inliers and outliers. $D_{in}$ performs the reconstruction for  inlier samples  while $D_{out}$ acts as a reconstructor for outliers. $D_{in}$ and $D_{out}$ work in a competitive way as they both try to give low reconstruction error for the input samples. However, each sample could only be interpreted by one decoder.
The architecture of the proposed model is shown in Fig.\ref{fig:arch}. It can be seen that both positive training data and unlabeled test data are feeded into ${E}$. As the training data are given as positive, we directly select $D_{in}$ to reconstruct them. For unlabeled test data, the label is determined by the reconstruction errors of two decoders.

Formally, let ${X}^p$ be the positive training data where ${X}^p_i$ is the $i$-th sample of  ${X}^p$, $i={1,...m}$. Similarly, let  ${X}^u$ be the unlabeled data and ${X}^u_j$ is the $j$-th sample of ${X}^u$, where $j={1,...,n}$. In order to assign labels for each ${X}^u_j$, we propose the following loss function.

\subsection{Competitive Reconstruction Loss}
Previous work~\cite{xia2015learning} showed that the reconstruction error of an auto-encoder, trained on samples from the target class, is a useful measure for novelty sample detection. Concretely, if an auto-encoder is trained to reconstruct target class samples (inliers), the reconstruction error for novelty samples would be high. In order to model the distributions of positive data and outliers separately. We design a competitive mechanism for our model. By the guidance of positive data, $D_{in}$ is trained to learn the distribution of target class. For unlabeled data which may be inliers or outliers, if the reconstruction error of $D_{in}$ is small than that of $D_{out}$, it could be an inlier with high probability. In this paper, we use mean square error as the reconstruction error for all our experiments. For simplicity, we define the reconstructions for sample $x$ as $\mathcal{R}_{in}(x)=D_{in}\Big(E(x)\Big)$ and $\mathcal{R}_{out}(x)=D_{out}\Big(E(x)\Big)$.

The competitive reconstruction loss is defined as
\begin{equation}
\begin{split}
\mathcal{L_{CR}} &= \left\| {X}^p -\mathcal{R}_{in}\Big({X}^p\Big)\right\|_2^2  \\
&+ \sum_{j=1}^n\bigg( y_j \left\| X^u_j - \mathcal{R}_{in}\Big(X^u_j\Big)\right\|_2^2 \\
& + (1-y_j)\left\| X^u_j - \mathcal{R}_{out}\Big((X^u_j\Big)\right\|_2^2\bigg)
\end{split}
\label{eq:crloss}
\end{equation}
where $y_j, j={1,...,n}$ is the label assignment for the $j$-th sample, and it is evaluated by
\begin{equation}
	y_j=
 	\begin{cases}
	1, & \left\| X^u_j- \mathcal{R}_{in}\Big(X^u_j\Big)\right\|_2^2 < \left\| X^u_j - \mathcal{R}_{out}\Big((X^u_j\Big)\right\|_2^2\\
	0,	& \text{otherwise}
	\end{cases}
\end{equation}
Note that $y_j$ is updated for each iteration. We will give detailed description on how to optimize the model later. We omit the average of the sum notation for clarity.

As mentioned before, \emph{CoRA} maps different data to different regions in the subspace. Thus, an intuitive idea is to preserve the data structures of positive data and abnormal data in the subspace, so that similar examples stay close in subspace. We propose a structure-preserving regularization term to improve the performance of our model.

\subsection{Subspace Structure-preserving Regularization}
In addition to competitive loss, we also propose a regularization term to keep the data structure in subspace. Subspace regularization has been explored in previous works~\cite{agarwal2010learning,liu2014unsupervised,yao2015semi}. Here we employ a block symmetric affinity matrix, as we use both positive data and unlabeled test data for training. Formally, we build the affinity matrix of the training samples as $W$:
\begin{equation}
W =
\begin{cases}
\text{exp}\left(\frac{-{D}(x_i, x_j)}{\epsilon^2}\right) & i\in \mathcal{N}_j\  \text{or} \ j \in \mathcal{N}_i\\
0 & \text{else}
\end{cases}
\label{eq:3}
\end{equation}
where $D(x_i,x_j)$ is the distance measure of the data. $\mathcal{N}_i$ is the neighborhood of the $i$-th data point. $\epsilon>0$ is the bandwidth parameter. In our method, $W$ can be formulated as
\begin{equation}
W = \left [\begin{matrix}
W_{pp} & W_{pu}\\
W_{up} & W_{uu}
\label{eq:w}
\end{matrix}\right ]
\end{equation}
where $W_{pp}$ and $W_{uu}$ indicate the affinity matrix among positive data and unlabeled data respectively, $W_{pu}$ and $W_{up}$ all represent the affinity matrix between positive data and unlabeled data. As for unlabeled data, we set $W_{uu}$ to a zero matrix in order to avoid subspace structure being destroyed by the uncertain structure of unlabeled data. In other words, we do not consider the manifold structure among unlabeled data due to the existence of uncertainty.

The manifold regularizer can be defined as
\begin{equation}
\mathcal{L_R} =  \sum_{i, j=1}^{n+m} {W}_{ij} \Big( E({ x}_i) - E({ x}_j) \Big)^2
\label{eq:5}
\end{equation}
where $E(x_i)$ denotes the encoder feature of sample $x_i$.
This regularization term can force the encoder to map similar examples into a neighborhood. Note that there may be a confusion: the sizes of positive data and unlabeled data are different, how could we formalize $W$ in Eq.~(\ref{eq:w}). It is a problem related to  the optimization algorithm, which will be explained in the next subsection.

\subsection{Optimization}
Competitive reconstruction auto-encoder is optimized to minimize the following loss function:
\begin{equation}
\mathcal{L}=\mathcal{L_{CR}}+\lambda\mathcal{L_R}
\end{equation}
where $\lambda>0$ is a hyperparameter that controls the relative importance of the regularization terms. We tuned $\lambda$ for different tasks and observed that $\lambda=0.1$ produces promising results for all our experiments. To optimize this loss function, stochastic gradient descent (SGD) method is adopted for training our model. As we use mini-batch SGD, for each iteration we sample the same number of positive data such that $W$ can be calculated as in Eq.~(\ref{eq:w}).
The detailed optimization algorithm is described in Alg.~\ref{alg:optimization}.
\begin{figure*}[htpb]
	\centering
	\includegraphics[width=0.96\textwidth]{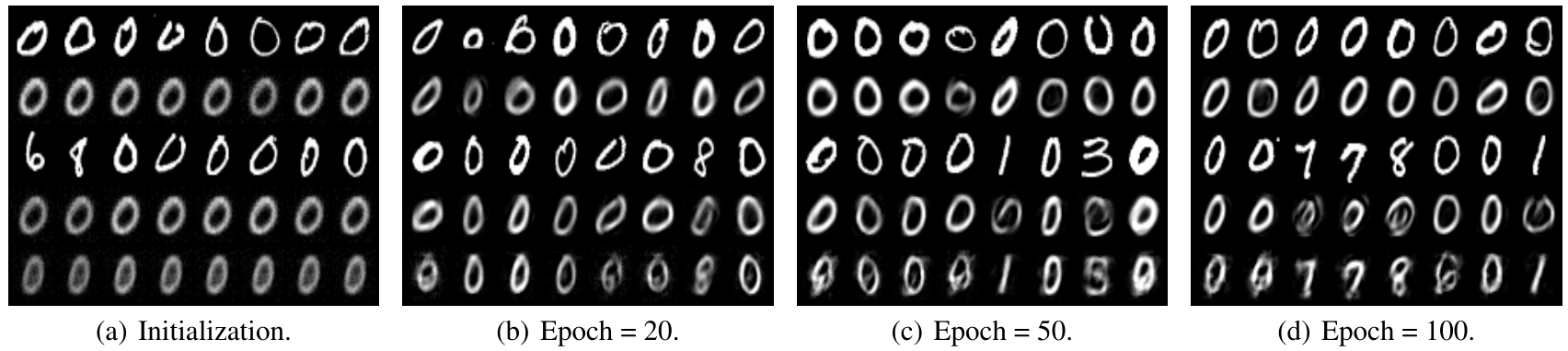}
	\caption{Reconstruction results of \emph{CoRA} after different training iterations. In each figure, samples in the 1st line are positive data, the 2nd line gives the reconstruction of the inlier decoder for the positive samples. Samples in the 3rd line are unlabeled data, the 4th line gives the reconstruction of the inlier decoder and the 5th line gives the reconstruction of the outlier decoder, all for the unlabeled samples.}
	\label{fig:envolve}
\end{figure*}

\begin{algorithm}[tb]
	\caption{Training Algorithm of \emph{CoRA}}
	\label{alg:optimization}
	\begin{algorithmic}
		\STATE {\bfseries Input:} positive data $X^p$, unlabeled test data $X^u$, encoder parameter $\theta_{E}$, decoder parameters $\theta_{\mathcal{D}_{in}},\theta_{\mathcal{D}_{out}}$
		\STATE Initialize parameters.
		\REPEAT
		
		\STATE Sample an unlabeled batch data $B^u$ and a positive batch data $B^p$, $|B^p|=|B^u|$.
		\STATE /* Forward propagation */
		\STATE $\hat{B}^p = \mathcal{D}_{in}\Big(E(B^p)\Big)$
		\STATE $\hat{B}^{iu}=\mathcal{D}_{in}\Big(E(B^u)\Big)$ /* Reconstruction  on $\mathcal{D}_{in}$*/
		\STATE$ \hat{B}^{ou} = \mathcal{D}_{out}\Big(E(B^u)\Big)$ /* Reconstruction  on $\mathcal{D}_{out}$*/

		\FOR{ each unlabeled sample $i=1$  to $|B^u|$}
		\IF{$\left\| \hat{B}^{iu}_{i} - {B}^u_i\right\|_2^2 <  \left\| \hat{B}^{ou}_i -  B^u_i\right\|_2^2$}
		\STATE Assign $B^u_i$ inlier decoder $\mathcal{D}_{in}$. %Accumulate$ \left\| \hat{B}^{iu}_{i} - {B}^u_i\right\|_2^2$ to $\mathcal{L_{CR}}$.
		\ELSE
		\STATE Assign $B^u_i$ outlier decoder $\mathcal{D}_{out}$. %Accumulate $\left\| \hat{B}^{ou}_i -  B^u_i\right\|_2^2$ to $\mathcal{L_{CR}}$.
		\ENDIF
		
		Compute  $\mathcal{L_{CR}}$ according to Eq.~\ref{eq:crloss}
		\ENDFOR
		\STATE Compute loss $\mathcal{L_R}$ with Eq.~\ref{eq:3},~\ref{eq:w},~\ref{eq:5}.
		\STATE Optimize $\mathcal{L}$ with SGD, and back propagate.
		\UNTIL{$convergence$}
	\end{algorithmic}
\end{algorithm}

\subsection{Discussions}
\textbf{The learning procedure}.
To understand the learning procedure of \emph{CoRA}, we explore the reconstruction results of inlier decoder and outlier decoder for different learning iterations. As shown in Fig.~\ref{fig:envolve}, in the very beginning of training, both decoders give similar reconstruction for different samples. However, after a few epochs, the inlier (positive) decoder begins to generate better reconstruction on target class than the outlier decoder. Finally, with the convergence of the learning procedure, the inlier decoder generates bad reconstruction for outliers which like distorted positive samples. Although the outlier decoder can output understandable reconstructions for positive samples, they all have some distortion. Thus, the positive samples are assigned to the inlier decoder, and outlier samples are assigned to outlier decoder.

\begin{figure*}[htpb]
	\centering
	\includegraphics[width=0.96\textwidth]{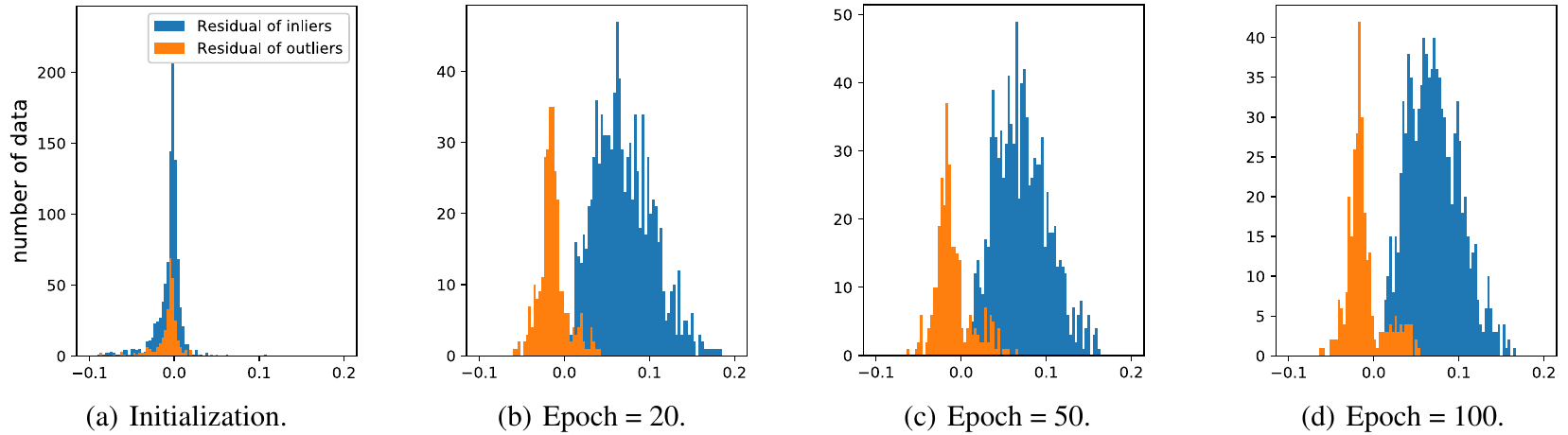}
	\caption{The evolution of reconstruction residual in the learning process of CoRA.}
	\label{fig:distribution}
\end{figure*}

Furthermore, we investigate the distributions of reconstruction residual between the inlier decoder and the outlier decoder on positive data/inliers and outliers during training. The residual is defined as $R_{residual}(x)=\left\| x - \mathcal{R}_{out}(x)\right\|_2^2 - \left\| x- \mathcal{R}_{in}(x)\right\|_2^2$. The results are shown in Fig.~\ref{fig:distribution}. We can see that both decoders have similar reconstruction error for inliers and outliers at the initialization stage. However, as the training proceeds, the residuals are different for inliers and outliers. After about 50 epochs, the learning process begins to converge and the discrimination between inliers and outliers becomes more and more clear. However, as there always exist some outliers that are hard to tell from inliers, thus there is a small overlap area.

\noindent\textbf{The merits of CoRA}.
As mentioned before, our model does not require a pre-defined threshold to discriminate outliers. This is quite important for many real-world applications. Another merit of our algorithm is that it does not suffer from the overfitting problem while most reconstruction-based models do. When the learning process converges, the assignment of each unlabeled sample rarely changes, which means CoRA is robust. Moreover, we do not impose any restriction on the design of the auto-encoder. By exploiting the representation learning ability of deep autoecoder, our model is flexible for various data. Meanwhile, mini-batch gradient descent optimization makes our approach efficient in handling large-scale datasets, which is a basic requirement in the big data era.

\section{Performance Evaluation}
In this section, we evaluate CoRA for anomaly detection task, and compare it with existing methods including the state of the art ones. We use seven datasets including low-dimensional data and image data. 
\subsection{Datasets}
\begin{itemize}
	\item\textbf{KDD99}. A classical anomaly detection dataset that records the network attacks and normal connections. We use a subset of it which is referred as \emph{KDD99 10-percent}. There are 494,021 instances, where 30\% are outliers.
	\item\textbf{MNIST}.
	It has 70,000 training and test samples from 10 digit classes. For each class, we select the training positive data and simulate outliers from the test data with different ratios, from $10\%$ to $50\%$.
	\item\textbf{Fashion MNIST}. The dataset composes of a training set of 60,000 examples and a test set of 10,000 examples. Each example is a $28\times28$ grayscale image, associated with a label from 10 classes, each of which is a cloth category. We preprocess it as for MNIST.
	\item\textbf{ImageNet-20}, This dataset consists of images in 20 semantic concepts from ImageNet dataset, and each concept contains about 2,800 images on average. For each concept, outliers are simulated from the other concepts with a ratio of $30\%$.
	\item\textbf{Caltech-101}.
	This dataset consists of 101 classes of images. Following previous work~\cite{zhai2016deep}, we choose the 11 object categories that contain at least 100 images. For each category, outliers are sampled from other categories with a ratio $30\%$.
	\item\textbf{CIFAR-10}.
	There are 50,000 training and 10,000 test images from 10 categories. Similar to Caltech-101, outliers are sampled from other categories at the ratio of $30\%$.
	\item\textbf{Caltech-256}. This dataset contains 256 object classes with a total of 30,607 images. Each category has at least 80 images. Following previous works~\cite{you2017provable,sabokrou2018adversarially}, we randomly select images from $n\in\{1,3,5\}$  categories as inliers, and for those categories that have more than 150 images, only the first 150 images are used. A certain number of outliers are randomly selected from the ``clutter'' category, such that each experiment has exactly 50\% outliers. We repeat each experiment three times and report the average performance.
	\end{itemize}
In most of our experiments, images are represented by features extracted from a pre-trained Vgg-16 network~\cite{simonyan2014very} on ImageNet. We use the features after the first fully connected layer and the feature dimension is 4096. On MNIST and Fashion-MNIST, we directly use raw pixels as inputs for all methods. As some approaches are restricted to simple multiple-layer neural networks, we use the same architecture for most of the networks for fair comparison.

\begin{figure}
	\centering
	\includegraphics[width=0.5\textwidth]{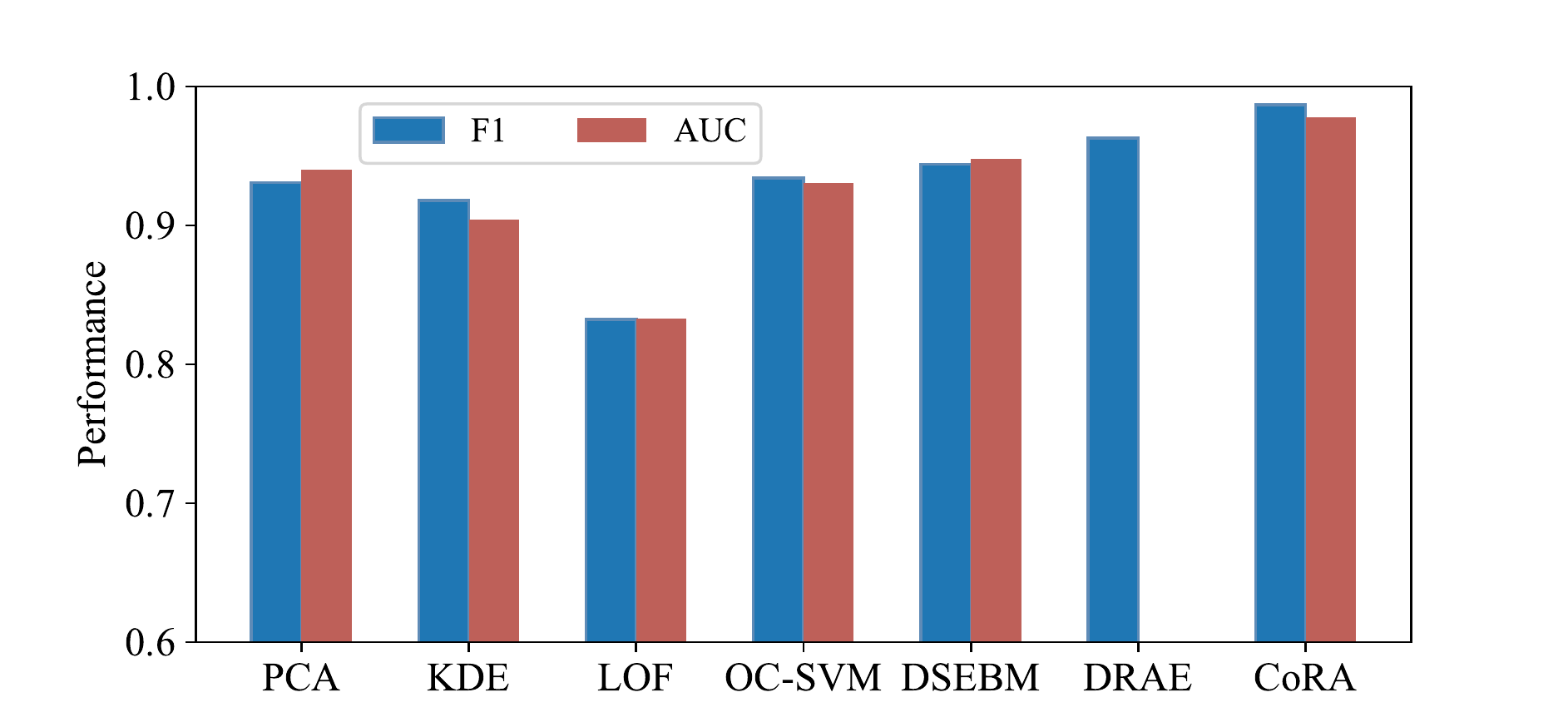}
	\caption{Performance comparison on KDD99 dataset.}
	\label{fig:kdd}
\end{figure}

\textbf{Evaluation Metrics}
Similar to previous works~\cite{sabokrou2018adversarially}, we utilize \emph{F1} and \emph{AUC} scores to measure performance. For most image datasets, we report the mean performance of all classes. Note that for discriminative algorithms such as DRAE, AUC may not suitable for evaluation. For our model, we calculate AUC with the transformed reconstruction residual in order to compare it with other supervised approaches.

\subsection{Experimental Results}
\subsubsection{Results on KDD99}
We first evaluate our \emph{CoRA} method on low-dimensional dataset. Since feature dimension of KDD99 is $41$ which is quite small, we compare \emph{CoRA} with several traditional methods that are effectiveness on low-dimension datasets, including PCA, KDE, LOF and OC-SVM. For deep learning models, we select deep energy-based model (DSEBM)~\cite{zhai2016deep} and discriminative auto-encoder (DRAE)~\cite{xia2015learning} for comparison.
The results are shown in Fig.~\ref{fig:kdd}. We can see that \emph{CoRA} achieves the best performance in terms of both \emph{F1} and \emph{AUC} scores. Although conventional methods such as PCA and OC-SVM perform comparably to deep neural network models including DSEBM and DRAE, the training of deep learning models is about 10 times faster than the conventional methods.

\begin{table}[!htpb]

	\caption{Performance on ImageNet-20, Caltech-101 and CIFAR-10. The best performance in each column is in \bf{bold}.}
	\label{tb:high}
	\small
	\begin{center}
		\begin{tabular}{c|cc|cc|cc}
			
			\hline \hline
			Dataset & \multicolumn{2}{c|}{ImageNet-20} & \multicolumn{2}{c|}{Caltech-101} & \multicolumn{2}{c|}{CIFAR-10} \\
			\hline
			Method & F1 & AUC & F1 & AUC& F1 & AUC \\
			\hline
			AE & 0.911 & 0.907 & 0.919 & 0.923 & 0.828 & 0.822  \\
			DSEBM & 0.926 & 0.923 & 0.923 & 0.892 & 0.867 & 0.872 \\
			DRAE & 0.930& - & \textbf{0.968} & - & 0.862 & - \\
			DAOC & 0.945 & 0.939& 0.956&0.944& 0.876& 0.869\\
			\emph{CoRA}& \textbf{0.960}& \textbf{0.964}& 0.954&  \textbf{0.985} & \textbf{0.910} & \textbf{0.897}  \\
			\hline
		\end{tabular}
	\end{center}
\end{table}

\begin{table*}[!htpb]
\caption{Performance comparison on Caltech-256. Inliers are images taken from one, three, or five randomly chosen categories, and outliers are randomly selected from category 257 --- clutter with a ratio of $50\%$. In each row the best result is in \textbf{bold} and the second best in \textit{italic} typeface.}
\begin{center}
\begin{tabular}{ccccccccccc}
\hline
  $\#$ categories &Measure& {\footnotesize CoP} &   {\footnotesize REAPER} &  {\footnotesize OutlierPursuit} &  {\footnotesize LRR} &  {\footnotesize DPCP} &  {\footnotesize R-graph} &   {\footnotesize SSGAN} & {\footnotesize DAOC} & {\footnotesize \emph{CoRA}}\\
\hline   \hline
1&AUC  & 0.905 & 0.816 & 0.837 & 0.907 & 0.783 &\textit{ 0.948} & - & 0.942 & \textbf{0.968}\\
1&$F1$ & 0.880 & 0.808 & 0.823 & 0.893 & 0.785 &  0.914 & \textbf{0.977} & 0.928 & \textit{0.967}\\
\hline \hline
3&AUC  & 0.676 & 0.796  & 0.788 & 0.479 & 0.798  & 0.929 &- & \textit{0.938} & \textbf{0.962}\\
3&$F1$  & 0.718 & 0.784 & 0.779  & 0.671 & 0.777 & 0.880 &  \textbf{0.963} & 0.913 & \textbf{0.963}\\
\hline \hline
5&AUC  & 0.487 & 0.657 & 0.629 & 0.337 & 0.676 & {0.913} & - & \textit{0.923} & \textbf{0.952}\\
5&$F1$  & 0.672  & 0.716 &  0.711  & 0.667 & 0.715 &  0.858 & \textit{0.945} & {0.905}& \textbf{0.950}\\
\hline
\end{tabular}
\end{center}
\label{tb:caltech256}
\end{table*}

\subsubsection{Results on MNIST/Fashion-MNIST}
\begin{figure}%[htpb]
		\centering
		\includegraphics[height=2.2in]{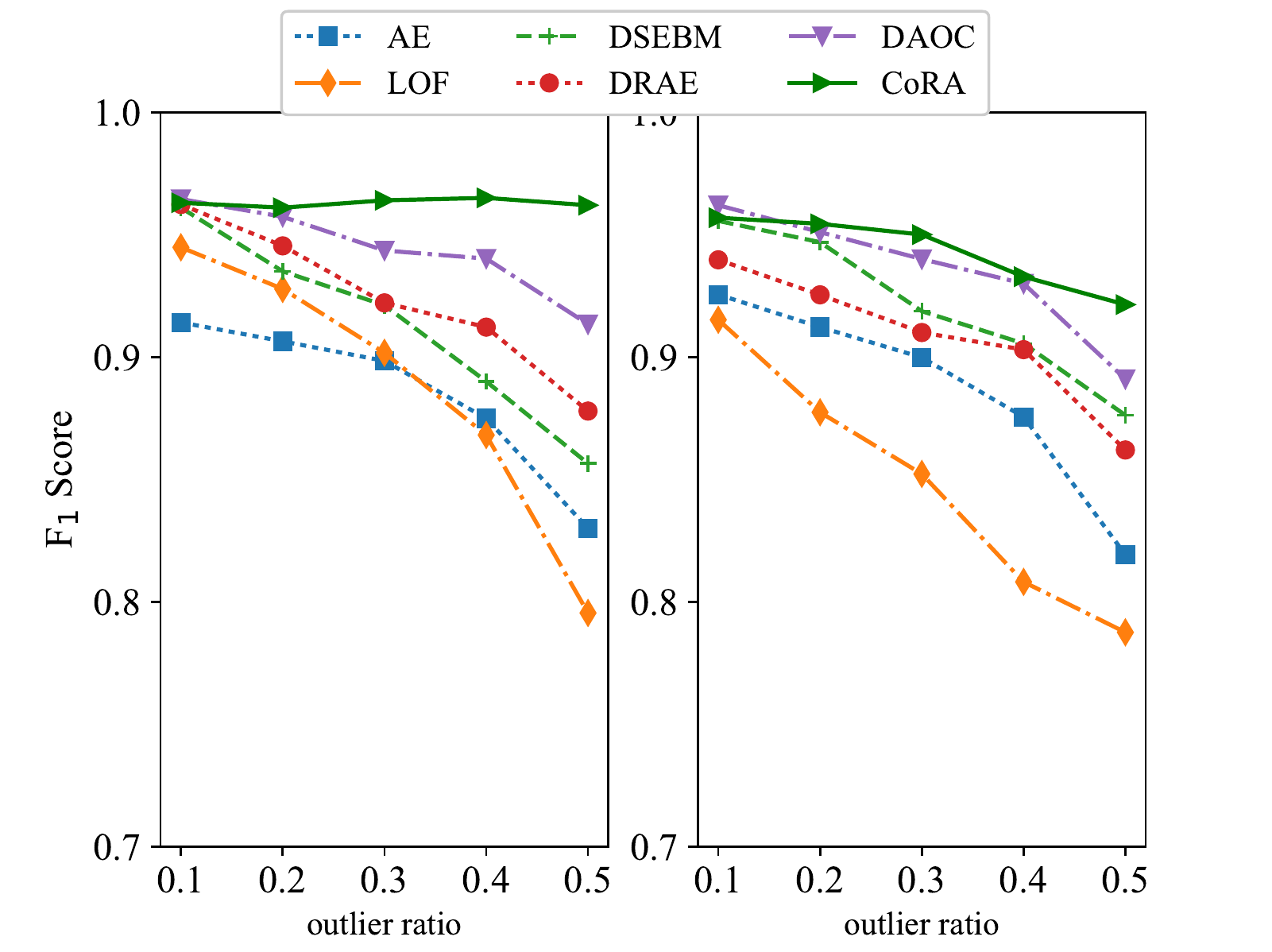}
	\caption{Performance comparison on MNIST (left) and Fashion-MNIST (right).}
	\label{fig:mnist}
\end{figure}

Raw pixels are used as input features for these two datasets, and the feature dimension is 784. In our experiments, the architecture of the encoder is $[784, 64, 32]$, the decoders use a symmetric architecture. As PCA is a special case of auto-encoder with a linear hidden layer, we omit PCA-based models in these experiments. We compare \emph{CoRA} with several state-of-the-art methods including DSEBM, DRAE and deep adversarial one-class learning (DAOC) method~\cite{sabokrou2018adversarially}. auto-encoder (AE) is used as the baseline algorithm. The results are shown in Fig.~\ref{fig:mnist}.

When the outlier ratio is small (\textit{e.g.} $0.1$), DAOC slightly outperforms CoRA on both datasets. The reason is that when the number of outliers is small, the outlier decoder of our model is not able to learn the anomaly data distribution well. However, CoRA still achieves comparable performance to unsupervised method DRAE and the other state-of-the-art algorithms, and it also outperforms the other traditional methods including AE.

With the increase of outlier ratio, most methods perform worse. However, CoRA shows its robustness to outlier ratio. As we can see, CoRA significantly outperforms all the other methods on MNIST dataset when outlier ratio is larger than $0.2$. Although similar trend can be observed on Fashion-MNIST, most state-of-the-art methods perform worse than on MNIST, since images in Fashion-MNIST are much more blurred than those in MNIST.

\subsubsection{Results on ImageNet-20/Caltech-101/CIFAR-10}
Except for DAOC~\cite{sabokrou2018adversarially}, most algorithms represent images by deep learning features. As DAOC uses auto-encoder as the generator and is trained in adversarial manner, we use the original pixel features as inputs to DAOC.
The architecture of auto-encoder for ImageNet-20 and CIFAR10 has two hidden layers with 256 and 128 units respectively. As there are too few samples in Caltech-101, we use a smaller architecture with two hidden layers. The encoder architecture is $[4096,64,32]$. We use \emph{ReLU} as the activation function of the hidden layers. Table.~\ref{tb:high} presents the results of state-of-the-art methods and AE. On ImageNet-20, DSEBM and DRAE achieve similar results while DAOC outperforms both of them.

CoRA performs best on ImageNet-20. For Caltech-101, DRAE achieves the best result in terms of \emph{F1} while CoRA achieves the best result in terms of \emph{AUC}. Meanwhile, CoRA and DAOC have close performance in terms of \emph{F1} measure. As mentioned before, the number of images in each concept of Caltech-101 is less than that in ImageNet-20 and CIFAR-10, it may not be sufficient for DAOC and CoRA to learn the inlier concept. As for CIFAR-10, DRAE slightly underperforms DSEBM, while DAOC outperforms the rest of approaches but CoRA. CoRA significantly outperforms the other methods in terms of both \emph{F1} and \emph{AUC}.

\subsection{Results on Caltech-256}
As there are less images in each category, detecting outliers from Caltech-256 is much challenging. Considering that Caltech-256 is a benchmark dataset for outlier detection task, in addition to DAOC, we compare our method with 7 other methods therein designed specifically for detecting outliers. Those methods include Coheerence Pursuit (CoP)~\cite{rahmani2017coherence}, OutlierPursuit~\cite{xu2010robust}, REAPER\cite{lerman2015robust}, Dual Principal Component Pursuit (DPCP)~\cite{tsakiris2015dual}, Low-Rank Representation (LRR)~\cite{liu2010robust}, OutRank~\cite{moonesignhe2006outlier}, and inductive semi-supervised GAN (SSGAN)~\cite{kimura2018semi}.

We use similar setup as in~\cite{you2017provable}, the experimental results are presented in Table~\ref{tb:caltech256}. The results of the other methods are from \cite{sabokrou2018adversarially} and \cite{kimura2018semi}. These results show that even for a small number of training samples, our method performs at least as well as the state-of-the-art algorithms, and in many cases it superior to them. Furthermore, when the inliers come from only one categories, CoRA slightly underperforms SSGAN. When the number of inlier classes increases to 3 or 5, CoRA outperforms the other 8 methods in terms of both $F1$ and $AUC$.

\section{Conclusion}
In this paper, we propose a competitive reconstruction auto-encoder model for transductive semi-supervised anomaly detection task. Our model learns from positive data and unlabeled test data, and predicts the labels of unlabelled data directly after learning. The two decoders are designed to compete with each other to achieve lower reconstruction error. With the guidance of positive data, inlier decoder is more likely to build the configuration of positive class. The new discriminative criterion does not need a pre-defined threshold, which is different from most existing methods. Moreover, we adopt SGD to optimize this model, which enables it to be efficient and scalable for large-scale datasets. Experimental results on seven benchmark datasets shown that this model can beat many state-of-the-art methods.

\noindent\textbf{Acknowledgement}: This work was supported by NSFC under gant No.~U1636205. Jihong Guan was supported by NSFC under grant No.~61772367.

\nocite{langley00}

\bibliographystyle{aaai}
\bibliography{crae}
\end{document}